\documentclass{elsarticle}
\usepackage{lineno,hyperref}
\modulolinenumbers[5]
\usepackage{booktabs,makecell}
\usepackage{multicol}
\usepackage{multirow}
\usepackage{amsmath,stmaryrd}
\usepackage{graphicx}
\usepackage{subfigure}
\usepackage{enumerate}
\usepackage{pgfplots}
\usepackage{tikz}
\usepackage{algorithm}
\usepackage[noend]{algpseudocode}

\usepackage{xcolor,colortbl}
\usepackage{footnote}
\usepackage{url}
\usepackage{array}
\usepackage{etex}

\journal{Journal of Pattern Recognition}

\begin{document}
	
	\begin{frontmatter}

		\title{DeepOtsu: Document Enhancement and Binarization using Iterative Deep Learning}
		
		\author{Sheng He\corref{cors}}
        \cortext[cors]{Corresponding author}
        \ead{heshengxgd@gmail.com}
        
        \author{Lambert Schomaker}
        \ead{L.Schomaker@ai.rug.nl}
        
        \address[author2]{Bernoulli Institute for Mathematics, Computer Science and Artificial Intelligence, University of Groningen, 
        	PO Box 407, 9700 AK, Groningen, The Netherlands}

		\begin{abstract}
		This paper presents a novel iterative deep learning framework and apply it for document enhancement and binarization. 
		Unlike the traditional methods which predict the binary label of each pixel on the input image, 
		we train the neural network to learn the degradations in document images and produce the uniform images of the degraded input images, which allows the network to refine the output iteratively. 
		Two different iterative methods have been studied in this paper: recurrent refinement (RR) which uses the same trained neural network in each iteration for document enhancement  and stacked refinement (SR) which uses a stack of different neural networks for iterative output refinement.
		Given the learned uniform and enhanced image, the binarization map can be easy to obtain by a global or local threshold.
		The experimental results on several public benchmark data sets show that our proposed methods provide a new clean version of the degraded image which is suitable for visualization
		and promising results of binarization using the global Otsu's threshold based on the enhanced images learned iteratively by the neural network.
			
		\end{abstract}
		
		
	\end{frontmatter}
%	\linenumbers
	
	\section{Introduction}
Extracting useful information from historical document images is a challenging problem because they usually suffer from different degradations~\cite{ntirogiannis2013performance}, such as noise, spots, bleed-through or low-contrast ink strokes~\cite{tonazzini2010color}.
A modern retrieval system, such as the Monk system~\cite{van2008handwritten} which is a web-based search engine in handwritten image collections, can only provide satisfying  results on high-quality handwritten images.
In addition, most methods for document analysis require preprocessed and clean documents as inputs to achieve a good performance~\cite{he2016multiple,stauffer2018keyword}.
Document enhancement and binarization is the main pre-processing step in document analysis process.
Document enhancement is the problem to improve the perceptual quality of document images and remove degradations and artifacts present in images~\cite{moghaddam2010variational}, aiming at restoring its original look~\cite{hedjam2013historical}.
  Document binarization is the task to separate each pixel to text and background~\cite{ntirogiannis2013performance}.
The enhancement is also a pre-processing step for binarization on degraded document images in order to remove some unnecessary noise.
Although many document enhancement methods have been proposed in the literature, most of them focus on a specific problem, such as the bleed-through correction~\cite{moghaddam2009low,moghaddam2010variational,yagoubi2015new,sun2016blind}.
Few existed unsupervised methods can handle various degradations in historical documents.

Convolutional Neural Networks (CNNs)~\cite{lecun1998gradient} has been successfully used in image classification~\cite{krizhevsky2012imagenet} and it provides significant improvements in various applications than traditional methods.
It has also been applied for document binarization~\cite{calvo2017pixel,vo2018binarization}.
Since the output of binarization has the same size of the input image, the famous frameworks of neural networks are often used, such as the fully convolutional neural networks (FCNs)~\cite{long2015fully,tensmeyer2017document}, holistically-nested edge detector (HED)~\cite{xie2017holistically,vo2018binarization} and U-Net~\cite{ronneberger2015u}.
Using deep learning provides a large margin of performance gain compared to traditional methods because millions of parameters in neural networks have been learned on a large training data set.

\begin{figure}
\centering
\includegraphics[width=0.8\textwidth]{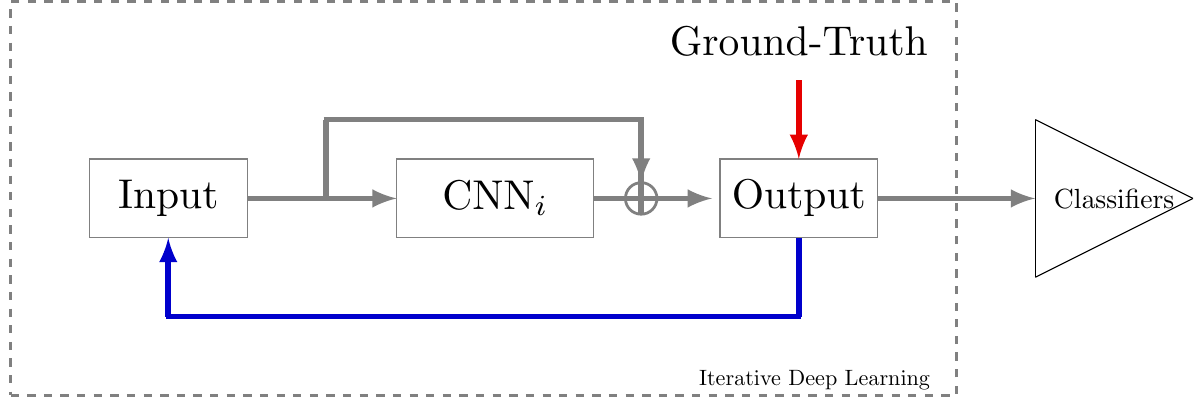}
\caption{The proposed iterative deep learning framework. The output of the CNN$_i$ is the modified version of the input, thus it can be fed into the network iteratively for fine-tunning with different iterations. The output after several iterations which is the improved version of the input can be considered as the input of the final classifiers (SVM or Neural networks).}
\label{fig:idlearning}
\end{figure}

In this paper, we propose the iterative deep learning framework, which is shown in Fig.~\ref{fig:idlearning}.
We train the network to improve the input images, such as removing noise or correcting some degradations.
Thus, the output of the neural network is the improved version of the input with supervised learning.
The neural network learns the differences between the input and expected output, which might be noises or other degradations.
The output can also be fed into the neural network for refinement with different iterations.
After several iterations, the output which is the improved version of the input, can be used as the input for the final classifiers to improve the performance.
The block of the iterative deep learning can be seamlessly integrated into any existed framework, which can be considered as the supervised data augmentation pre-processing.

In this paper, we apply the iterative deep learning for document enhancement and the final classifier is traditional binarization method.
Unlike the traditional binarization methods which train the neural network to predict the label of each pixel, we train the neural network to learn the degradation and correct the degraded document iteratively.
The output of the neural network is the expected uniform and clean images instead of the binary maps, which allows the network to learn the degradations: the differences between the degraded and clean images.
Since the output of the neural network is the improved version of the input, it can be also fed into the network for fine-tunning. 
More precisely, the learned neural network can be used recursively to refine the results because the output image can also be considered as 
a lightly degraded image if the learned neural network does not provide a good result in the first iteration.
In addition, given the uniform image which is corrected by the learned neural network, the binarized image can be easily and efficiently obtained by a global threshold, such as the Otsu's threshold~\cite{otsu1979threshold}.

\begin{figure*}[!t]
\centering
\includegraphics[width=0.9\textwidth]{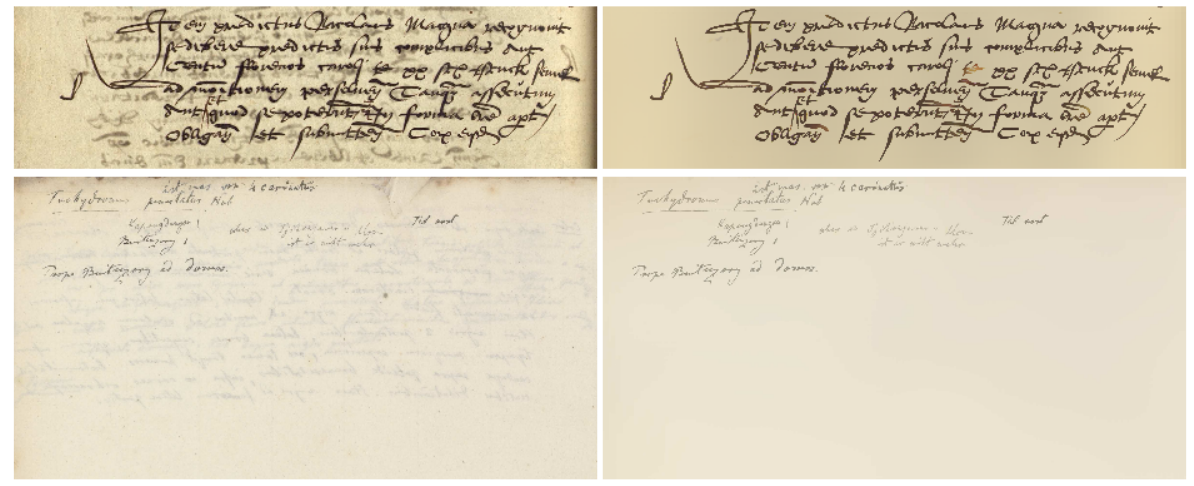}
\caption{Demonstration of document image enhancement on two images from the Monk system~\cite{van2008handwritten}. The first column shows the original degraded images and the second column shows the corresponding enhanced images of the proposed method.}
\label{fig:visualExamples}
\end{figure*}

Note that the output of the proposed method is the uniform and clean version of the degraded input image, which is an acceptable view of the degraded images for the end users, such as historian, paleographers and scholars.
The acceptable review is the visualization that the enhanced image should maintain the original appearance as much as possible while remove the textures and degradations on the background.
Our proposed method can provide a better view of the degraded document images which only shows the original text and the noise and degradations are removed.
Fig.~\ref{fig:visualExamples} presents two examples of the original degraded images and the corresponding enhanced images of the proposed method, which shows that the enhanced images are more readable for end users than the original documents.

In summary, the differences of the proposed method with the existed works~\cite{tensmeyer2017document,peng2017using,vo2018binarization,calvo2018selectional} are summarized as follows.
(1) Unlike the previous methods which train the neural network to learn the labels of each pixel,  the output of our method is the latent uniform version of the input images, which represents an internally enhanced version of the image.
(2) Our method can be used iteratively to refine the outputs since the output of the method is the improved version of the input. However, the previous methods are based on intensity probabilities per pixel, which are hard to optimize iteratively.
(3) In our approach we make a distinction between handling degradations and the handling of the binarization. 
The neural network is trained to correct degradations, while the final binarization is achieved by the efficient global-threshold Otsu method.

The rest of this paper is organized as follows: 
Section~\ref{sec:relatedwork} provides a short brief summary of a selection of related works.
The proposed method is present in Section~\ref{sec:proposedmethod} and the experimental results is reported in Section~\ref{sec:exps}.
Finally, Section~\ref{sec:conclusion} provides our conclusions and prospects for future works.

\section{Related Work}
\label{sec:relatedwork}
Binarization is a classical research problem for document analysis and many document binarization methods have been proposed over the past two decades in the literature.
It aims to convert each pixel in a document image into either text or background.
The most popular and simple method is the Otsu~\cite{otsu1979threshold}, which is a nonparametric and unsupervised method of automatic threshold selection approach for gray-scale image binarization.
It selects the global threshold based on the gray-scale histogram without any priori knowledge thus the computational complexity is linear.
The Otsu method works very well on uniform and clean images while produces poor results on degraded document images with nonuniform background.
In order to solve this problem, local adaptive threshold methods have been proposed, such as Sauvola~\cite{sauvola2000adaptive}, Niblack~\cite{niblack1986introduction}, Pai~\cite{pai2010adaptive} and AdOtsu~\cite{moghaddam2010multi,moghaddam2012adotsu}. 
These methods compute the local threshold for each pixel based on the local statistic information, such as the mean and standard deviation of a local area around the pixel.
It should be noted that binarization is not always the goal.
Methods such as Otsu can also be used for strong contrast enhancement.

Although the global or local threshold methods mentioned above are very efficient, their results are still not satisfied on high degraded and poor quality document images.
Therefore, document enhancement methods are usually used as preprocessing in order to remove degradations or noise in document images.
Several image processing techniques, such as the mathematical morphological operator and region-growing method are used in~\cite{shi2011image} for document enhancement and binarization.
Gatos et al.~\cite{gatos2006adaptive} use a Wiener filter to estimate the background surface which is involved in the final threshold computation.
Similarly, in~\cite{vo2018robust}, the background surface is estimated by a robust regression method and the document is binarized by a global thresholding operation.
Su et al.~\cite{su2013robust} propose an adaptive contrast map for text edge detection and the local threshold is estimated based on the mean and standard deviation of pixel values on the detected edges in a local region.
Nafchi et al.~\cite{nafchi2014phase} introduce a robust phase-based binarization method which involves image denoising with phase preservation. 
For bleed-through correction on degraded documents, a new variational model is introduced in~\cite{moghaddam2010variational} based on wavelet shrinkage or a time-stepping scheme.
A patch-based non-local restoration and reconstruction method is proposed in~\cite{moghaddam2011beyond} for degraded document enhancement.
In~\cite{sun2016blind}, a new conditional random field (CRF)-based method~\cite{song2015efficient}
is proposed to remove the bleed-through from the degraded images.
The bio-inspired model by the off-center ganglion cells of human vision system is used in~\cite{zagoris2017bio} for document enhancement and binarization.
All methods mentioned above use traditional techniques for document enhancement and each of them can only handle a certain type of degradation in document images.

Other priori knowledge of text is also exploit for binarization, such as the edge pixels extracted by edge detectors.
For example, the Canny edge detector is used to extract edge pixels in~\cite{chen2008double} and then the closed image edges are considered as seeds to find the text region.
The transition pixel which is a generation of the edge pixel is introduced in~\cite{ramirez2010transition} is computed based on the intensity differences in a small neighbor regions and the statistic information of these pixels are used to compute the threshold.
In~\cite{jia2018degraded}, structural symmetric pixels around strokes are used to compute the local threshold. 
Howe~\cite{howe2013document} proposes a promising method which can tune the parameters automatically with a global energy function as a loss which incorporates edge discontinuities (Canny detector is used).

Convolutional neural networks achieve good performance on various applications, which is also applied in document analysis.
For example, the winner of the recent DIBCO event~\cite{pratikakis2017icdar2017} uses the U-Net convolutional
network architecture for accurate pixel classification.
In~\cite{tensmeyer2017document}, the fully convolutional neural network is applied at multiple image scales.
The deep encoder-decoder architecture is used for binarization in~\cite{peng2017using,calvo2018selectional}.
A hierarchical deep supervised network is proposed in~\cite{vo2018binarization} for document binarization, which achieves start-of-the-art performance on several benchmark data sets.
In~\cite{westphal2018document}, the Grid Long Short-Term Memory (Grid LSTM) network is used for binarization.
However, it achieves lower performance than Vo's method~\cite{vo2018binarization}.

\section{Proposed Method}
\label{sec:proposedmethod}
In this section, the problem of document enhancement is discussed based on the iterative deep learning.
We first present the formulation of learning degradations for document enhancement and then the structure of CNN which is used for evaluation of the proposed model is introduced.

\subsection{Problem formulation}
\begin{figure}[!t]
	\centering 
	\includegraphics[width=0.8\textwidth]{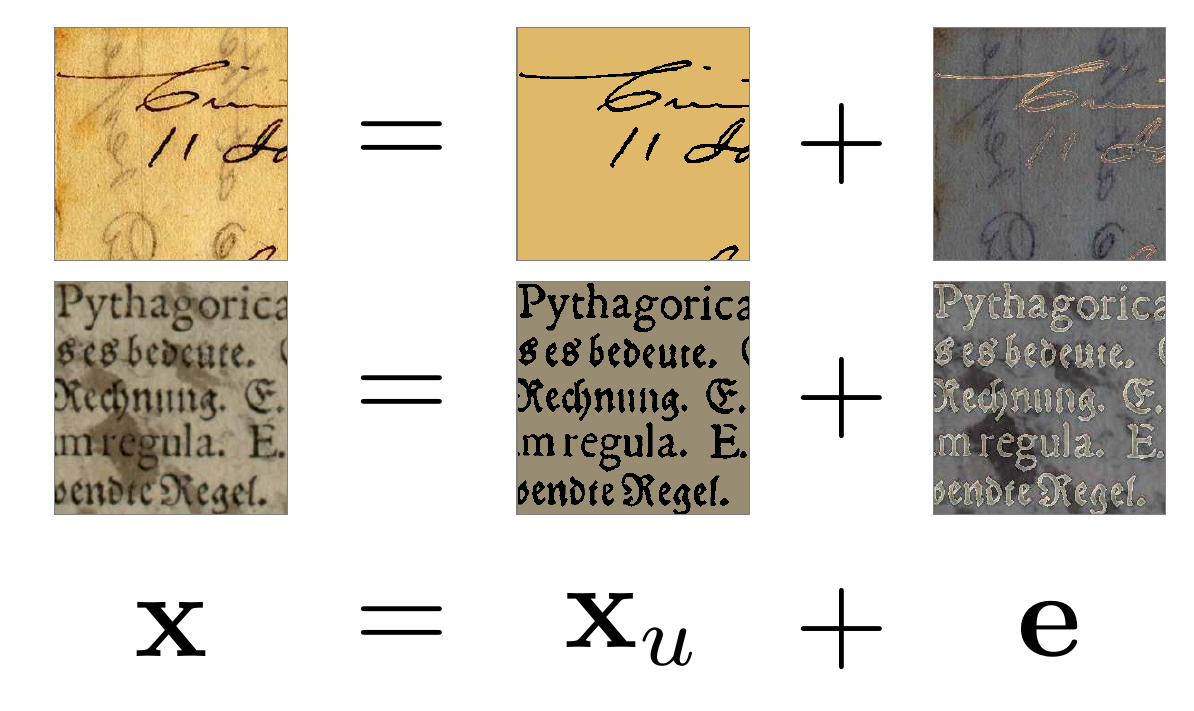}
	\caption{Schematic description of the proposed degradation model.
	 A degraded pattern $\textbf{x}$ in the degraded image is assumed to be the sum of an ideal (uniform) pattern $\textbf{x}_u$ and the degradation $\textbf{e}$.}
	\label{fig:model}
\end{figure}

An original clean or uniform document (ground-truth) is assumed to be degraded by various degradations, such as bleed-through or other artifacts.
In the image enhancement formulation, the value of each pixel in the degraded images is supposed to the sum of the original value and the degraded value, which can be expressed by:
\begin{equation}
\textbf{x} = \textbf{x}_{u} + \textbf{e}
\end{equation}
where $\textbf{x}$ is the degraded image, $\textbf{x}_u$ is the latent clean or uniform image and $\textbf{e}$ is the degradation.
The probability density of the $\textbf{e}$ depends on the type of degradations.
Fig.~\ref{fig:model} gives a visual example of this model.

Most methods for historical document analysis requires a clean or uniform image $\textbf{x}_{u}$ as input to extract the text edges or contours.
Therefore, recovering the clean image $\textbf{x}_{u}$ given the degraded image $\textbf{x}$ is a classical document enhancement problem.
When the uniform $\textbf{x}_{u}$ is available, document binarization is quite simple, which can be computed by a global threshold, such as Otsu~\cite{otsu1979threshold}.

The Convolutional Neural Network (CNN)~\cite{lecun1998gradient} has been successfully used in image classification~\cite{krizhevsky2012imagenet}, but it can be applied to document binarization~\cite{calvo2017pixel,vo2018binarization}. 
However, the traditional methods directly apply the CNN on the degraded image $\textbf{x}$ to compute the binary image $\textbf{x}_b$ by:
\begin{equation}
\label{eq:cnn}
\textbf{x}_b = \text{CNN}(\textbf{x})
\end{equation}
which in fact requires the CNN to implicitly learn the latent uniform image $\textbf{x}_{u}$, the degradation $\textbf{e}$ and also the threshold on the latent uniform image $\textbf{x}_{u}$.

In this paper, we train the neural network to predict the uniform image of the input by:
\begin{equation}
\label{eq:refunc}
\begin{split}
\textbf{x}_u & = \text{CNN}(\textbf{x}) + \textbf{x} \\
%\textbf{x}_b & = \text{Otsu}(\textbf{x}_u)
\end{split}
\end{equation}
here the function $\text{CNN}(\textbf{x}) = -(\textbf{x}-\textbf{x}_u)=-\textbf{e}$ represents the degradations (negative): the difference between the degraded and clean images, which is learned by the neural network.
If the input $\textbf{x}$ is a uniform and clean image, the neural network does not need learn any information and the output of the neural network is closed to zero.
Since the output $\textbf{x}_u$ is the improved version of the input $\textbf{x}$,
it is possible to improve the output $\textbf{x}_u$ iteratively by the neural network if we set $\textbf{x}=\textbf{x}_u$ in the next iteration.

When the the uniform $\textbf{x}_u$ image is obtained after several iterations, the binarization map can be computed by:
\begin{equation}
\label{eq:bin}
\textbf{x}_b = \mathcal{B}(\textbf{x}_u)
\end{equation}
where $\mathcal{B}$ can be any existed binarization methods, or learned neural networks.
Because the $\textbf{x}_u$ is the enhanced and clean image, the simple and efficient method can be used for binarization, such as the global Otsu's threshold~\cite{otsu1979threshold}.

Our new reformulation proposed in Eqn.~\ref{eq:refunc} is motivated by the residual learning~\cite{he2016deep}. 
However, the proposed method applies the CNN directly on the input image, which allows the neural network to learn degradations and correct the degraded images iteratively.
The CNN structure in Eqn.~\ref{eq:refunc} can be any neural networks, including the residual network~\cite{he2016deep}.

The advantages of the proposed model are:
(1) The neural network only learns the degradation of the image, without fitting the latent uniform images, such as the distribution of the background.
(2) The proposed method has a new intermediate output $\textbf{x}_u$, which can be considered as the enhanced image version of the degraded input $\textbf{x}$.
(3) This can be seamlessly integrated with other methods by Eqn.~\ref{eq:bin}.
For example, any existing binarization method can be applied on the estimated uniform image $\textbf{x}_u$ learned by the neural network.

The proposed model learns the uniform image $\textbf{x}_u$ directly from the original image.
However, the learned $\textbf{x}_u$ might not be perfect and it can also be considered as the degraded images $\textbf{x}$ if the network does not provide good results.
Thus, the learned $\textbf{x}_u$ can be refined or enhanced recursively if we set $\textbf{x}=\textbf{x}_u$ to the neural network.

\begin{figure}[!]
	\centering
	\includegraphics[width=0.8\textwidth]{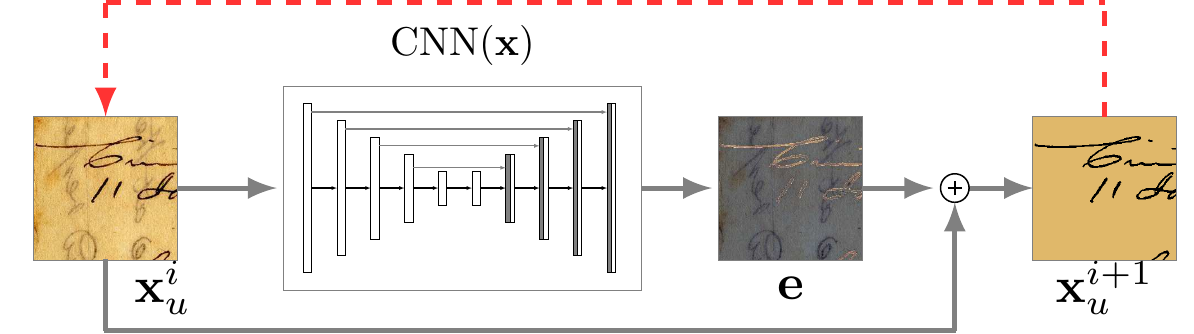}
	\caption{The recurrent refinement (RR) diagram of the $i$-th iteration. The red dashed line denotes that the output of the neural network can be used as input for iterative fine tuning with different iterations.
	$\textbf{x}^0_u=\textbf{x}$ is the original degraded image at the beginning when $i=0$.}
	\label{fig:offline}
\end{figure}

If we obtain the $\textbf{x}_u$ from the neural network, there are two ways to refine it iteratively: (1) feed it into the same neural network for fine enhancement, called ``Recurrent Refinement (RR)" and (2) train a new network (with the same or different structures), called `` Stacked Refinement (SR)".
The recurrent refinement (RR) method is defined as:
\begin{equation}
\label{eq:reci}
\begin{split}
\textbf{x}^i_u & = \text{CNN}(\textbf{x}^{i-1}_u) + \textbf{x}^{i-1}_u \\
\end{split}
\end{equation}
where $\textbf{x}^i_u$ is the $i$-th output of the neural network and $\textbf{x}^0_u=\textbf{x}$ which is the original degraded image. 
Note that there is only one neural network which is trained to refine the results iteratively.
Fig.~\ref{fig:offline} shows the diagram of the RR framework.
The advantage of the RR method is that once the neural network is trained, it can be used iteratively with many iterations.
However, this also requires the neural network to learn different levels of degradations in document images.
For example, it needs to remove noise on the background and recover the weak ink trace on the text region by the same network.

\begin{figure*}[!t]
	\centering
	\includegraphics[width=\textwidth]{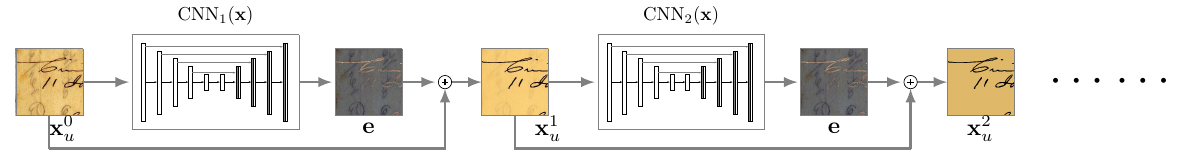}
	\caption{The stacked refinement (SR) diagram. The neural networks are stacked together to refine the results. Note that different neural networks with the same structure (CNN$_1$, CNN$_2$,...) are trained in this example.}
	\label{fig:online}
\end{figure*}

The stacked refinement (SR) method is defined as:

\begin{equation}
\begin{split}
\textbf{x}^i_u & = \text{CNN}_i(\textbf{x}^{i-1}_u) + \textbf{x}^{i-1}_u \\
\end{split}
\end{equation}
which is similar as the Eqn.~\ref{eq:reci}, but the new network $\text{CNN}_i$ is trained during the $i$-th iteration to refine the input $\textbf{x}^{i-1}_u$ which is the output of the $(i-1)$-th iteration on the $\text{CNN}_{i-1}$ neural network.
Fig.~\ref{fig:online} gives the an example of two stacked neural networks with the same structure.
However, the neural network structure can also be different in different iterations.
The SR method is better than the RR method because new network is trained iteratively to learn the degradations in different levels.
For example, the neural network in the first iteration can learn the distribution of the background and in the second iteration can learn the distribution of the ink traces.
Therefore, background noise removing and ink trace recovering can be performed in different networks.

Ideally, the RR and SR methods can be used in a mixed way.
For example, the neural network of the RR method can also be a stack of networks used in the SR method which is trained iteratively.
However, due to time and memory costs, we evaluate the proposed RR and SR methods separately in this paper.

\subsection{Network Architectures}
Although any neural networks can be used, in this paper we adopt the basic U-Net~\cite{ronneberger2015u} to learn degradations in historical documents,
similar to image segmentation.
The architecture consists two paths: contracting path and expansive path.
In the contracting path, there are five convolutional layers with the kernel size 3$\times$3, each followed by a leaky-ReLU~\cite{maas2013rectifier} ($\lambda=0.25$) and 2$\times$2 max pooling layer with stride 2.
In the expansive path, the deconvoluational operation is used to upsample the feature maps and then it is concatenated with the corresponding feature maps on the contracting path, followed by a convolutional layers.
The filter numbers in five convolutional layers are set to: [16,32,64,128,256], respectively.
The output layer has the same size as the input image, which makes the additive operation possible (shown in Eqn.~\ref{eq:refunc}).

The network is trained by minimizing the following loss function in each iteration:
\begin{eqnarray}
\label{eq:sloss}
L^i=\frac{1}{n}\sum |\textbf{x}_t-\hat{\textbf{x}}_u^i|
\end{eqnarray}
where $n$ is the number of pixels in the image, $\textbf{x}_t$ is the ground-truth and $\hat{\textbf{x}}_u^i$ is the prediction from the neural network with the input $\textbf{x}_u^i$ in the $i$-th iteration ($\textbf{x}_u^0=\textbf{x}$ which is the original degraded image at the beginning).
The network in each iteration is trained with the loss defined in Eqn.~\ref{eq:sloss} with the degraded image and the uniform ground truth $\textbf{x}_t$.
The network of the RR and sR models on each iteration can be trained separately and jointly.
In this paper, we train the network jointly and the combined loss is defined as
\begin{eqnarray}
L_{total}=\frac{1}{m}\sum_i L^i
\end{eqnarray}
where $m$ is the number of iterations and $L^i$ is the loss on the $i$ the iteration, which is defined in Eqn.~\ref{eq:sloss}.

\section{Experiments}
\label{sec:exps}
In this section, we present the experimental results of the proposed methods for document enhancement and binarization.
The training data sets are constructed based on several public benchmark data sets.
We also introduce a new bleed-through data set, called Monk Cuper Set (MSC) where the historical documents are collected from the Cuper book collection of the Monk system~\cite{van2008handwritten}.

\subsection{Dataset}
There are several public data sets for document binarization, such as (H-)DIBCO data sets which are used for document binarization competition. 
Similar as~\cite{vo2018binarization}, we select images on DIBCO 2009~\cite{gatos2009icdar}, H-DIBCO 2010~\cite{pratikakis2010h} and H-DIBCO 2012~\cite{pratikakis2012icfhr}  for training.
The training set also includes documents on the Bickely-diary dataset~\cite{deng2010binarizationshop}, PHIDB~\cite{nafchi2013efficient} and the Synchromedia Multispectral dataset~\cite{hedjam2015icdar}.
The documents on DIBCO 2011~\cite{pratikakis2011icdar},DIBCO 2013~\cite{pratikakis2013icdar}, H-DIBCO 2014~\cite{ntirogiannis2014icfhr2014} and H-DIBCO 2016~\cite{pratikakis2016icfhr2016} are selected for evaluation.
Four evaluation metrics which are used in the (H-)DIBCO contests, are adopted in this section for quantitatively evaluation and comparison, including F-measure, pseudo F-measure ($F_{ps}$), distance reciprocal distortion metric (DRD) and the peak signal-to-noise ratio (PSNR).

Following contest reports~\cite{gatos2009icdar,pratikakis2010h,pratikakis2012icfhr,pratikakis2013icdar,ntirogiannis2014icfhr2014,pratikakis2016icfhr2016}, these evaluation metrics are defined as follows:

\begin{enumerate}
	\item F-measure (FM):
	\begin{equation}
		FM = \frac{2\times Recall \times Precision}{Recall + Precision}
	\end{equation}
	where $Recall=\frac{TP}{TP+FN}, Precision=\frac{TP}{TP+FP}$, $TP,FP,FN$ denote the True positive, False position and False Negative values, respectively.
	\item pseudo F-measure ($F_{ps}$):
	\begin{equation}
		F_{ps} = \frac{2\times pRecall \times Precision}{pRecall + Precision}
	\end{equation}
	where $pRecall$ is the percentage of the skeletonized ground truth image described in~\cite{pratikakis2010h}.
	\item distance reciprocal distortion metric (DRD):
	\begin{equation}
		DRD = \frac{\sum_k DRD_k}{NUBN}
	\end{equation}
	where $DRD_k$ is the distortion of the $k$-th flipped pixel and it is calculated using a 5$\times$5 normalized weight matrix and $NUBN$ is the number of the non-uniform 8$\times$8 blocks in the ground truth image (see details in~\cite{ntirogiannis2014icfhr2014}).
	\item peak signal-to-noise ratio (PSNR):
	\begin{equation}
		PSNR = 10\text{log}(\frac{C^2}{MSE})
	\end{equation}
	where $MSE=\frac{\sum_{x=1}^M\sum_{y=1}^N(I_{bin}(x,y)-\hat{I}_{bin}(x,y))^2}{MN}$, $C$ denotes the difference between the text and background.
\end{enumerate}

We also construct a new data set for evaluation, Monk Cuper Set (MCS), which contains 25 pages sampled from a real historical collections.
The documents in this set have a heavy bleed-through degradations and textural background, making them very hard for information retrieval by a computer and even hard for end users to read.
Several examples are shown in Fig.~\ref{fig:mcsEnhanceExamp}.
This data set is available on the author's website for academic usage.

\subsection{Implement details}

\textbf{Data preparation}. We train our networks with small image patches sampled from the document images with a sliding window. 
The basic patch size is set to 256$\times$256 (suggested in~\cite{calvo2018selectional}).
Data augmentation is very important for boosting the performance of the neural network and we also apply augmentation methods (scale and rotation) for creating more training samples.
For scale augmentation, we sample patches with the scale factor \{0.75,1.25,1.5\} based on the input of the neural network and resize them to  256$\times$256.
For rotation augmentation, we rotate each patch with a rotation angle 270.
Overall, more than 120,000 training patches are created for training.

\begin{figure}[!t]
\centering 
\includegraphics[width=\textwidth]{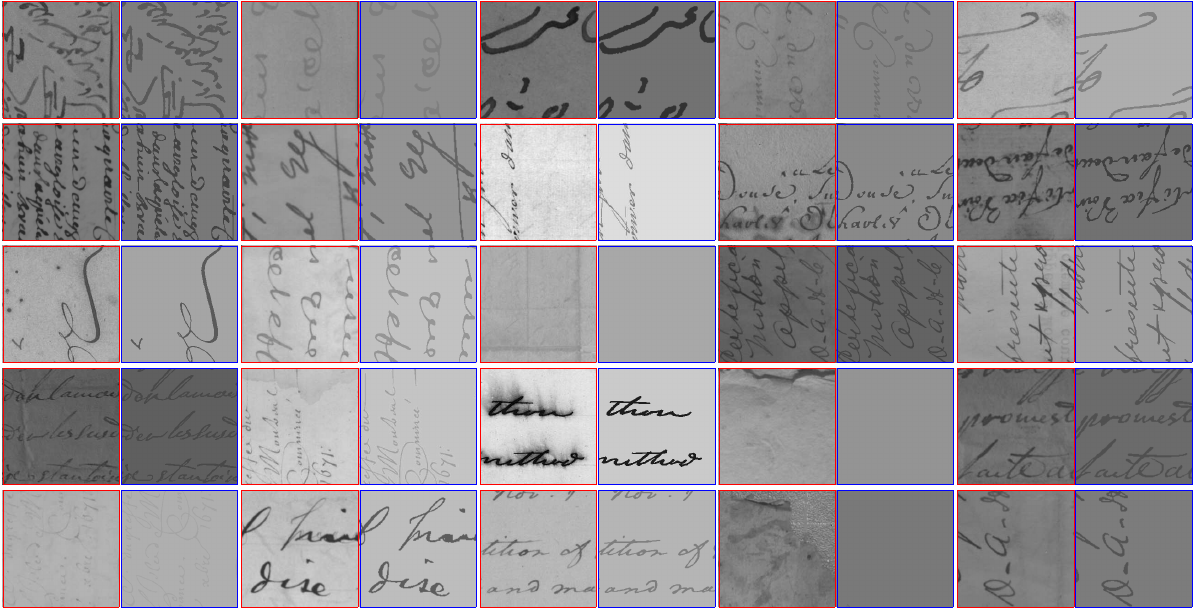}
\caption{Training samples (red box) with their corresponding ground-truth (blue box) images.
Each pixel in the ground-truth is the average of pixels with the same label (text or background) within the patch.}
\label{fig:gtSample}
\end{figure}

\textbf{Ground-truth construction}. Since the output of the neural network is the uniform images of the input, each pixel value on the ground-truth image is computed as the average pixel value with the same label within the patch. 
The text and background label is obtained from the ground-truth of the binary maps.
For example, for the patches which do not contain any text strokes or ink traces, the ground-truth is the average image of the patch, which is helpful to remove noise in the background regions of document images.
Fig.~\ref{fig:gtSample} shows the training samples used in this paper.

\textbf{Training}. The training batch size is set to 5 due to the limitation of the memory.
The learning rate is set to $10^{-4}$ and the number of training iteration is 110,000.
The system runs on a PC platform with a single GPU (NVIDIA GTX 960 with 4G memory).

\subsection{Document enhancement}
\begin{figure*}[!t]
\centering 
\includegraphics[width=\textwidth]{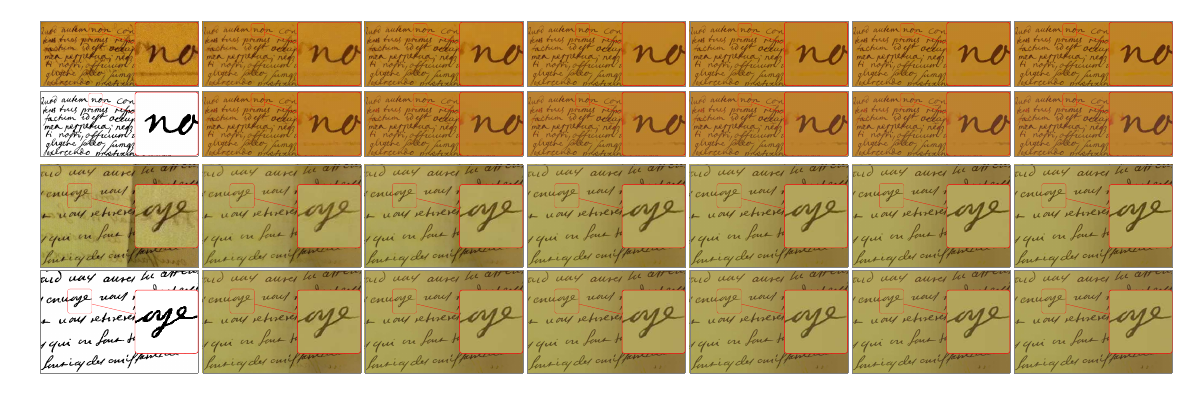}
\caption{Examples of enhancement with different iterations of the SR (top row) and RR (bottom row) methods.
The first column shows the original images (top) and the corresponding binary maps (bottom row). Images from second column to the last are the corresponding results of $i$-iteration where $i=1,2,\cdots,6$. }
\label{fig:iterenhan}
\end{figure*}

Given an image, the patches with the same size as training patches are sampled with a sliding window strategy.
The values of each pixel in the enhanced image are the average values of the overlapping patches computed from the trained neural networks.
Fig.~\ref{fig:iterenhan} shows a visual example of two documents (in RGB color space) on the DIBCO 2013 data set with different iterations by the SR and RR methods.
It shows that with more iterations, the bleed-through and noise on background are removed and the ink traces are enhanced.

\begin{figure*}[!t]
\centering 
\includegraphics[width=\textwidth]{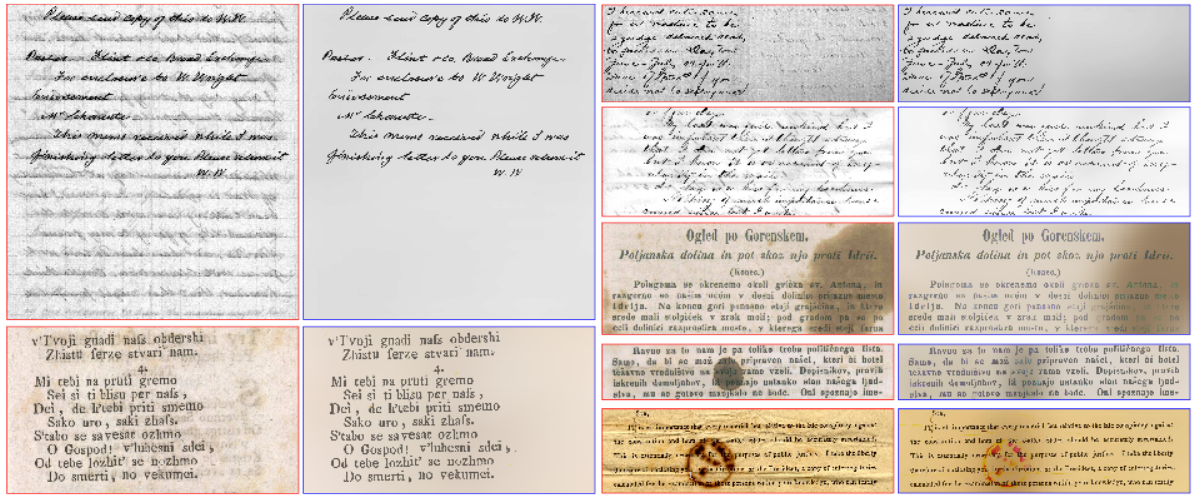}
\caption{Examples of enhancement document on the DIBCO2013 data set by the SR method.}
\label{fig:dibc02013examp}
\end{figure*}

\begin{figure*}[!t]
\centering 
\includegraphics[width=\textwidth]{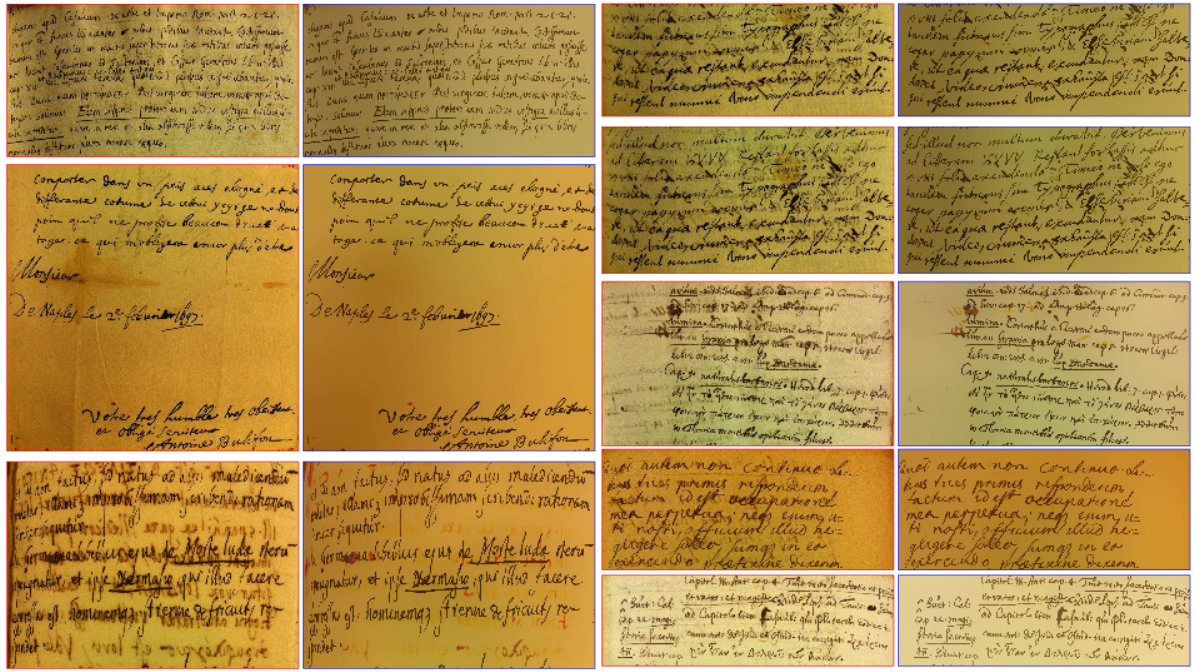}
\caption{Examples of enhancement document on the challenging MCS data set by the SR method.}
\label{fig:mcsEnhanceExamp}
\end{figure*}

Fig.~\ref{fig:dibc02013examp} and~\ref{fig:mcsEnhanceExamp} show results of the enhancement documents computed by the SR method on the DIBCO 2013 and MCS data sets, respectively.
From the figure we can see that: (1) The outputs are very clean. The noise and bleed-through degradations on the background are removed.
(2) The large smears in the document images can not be removed completely but smoothed, because the input of the neural network is a small patch and the ground-truth of the proposed method is constructed based on the small patch, instead of the global images.

\begin{figure*}[!t]
\centering 
\includegraphics[width=\textwidth]{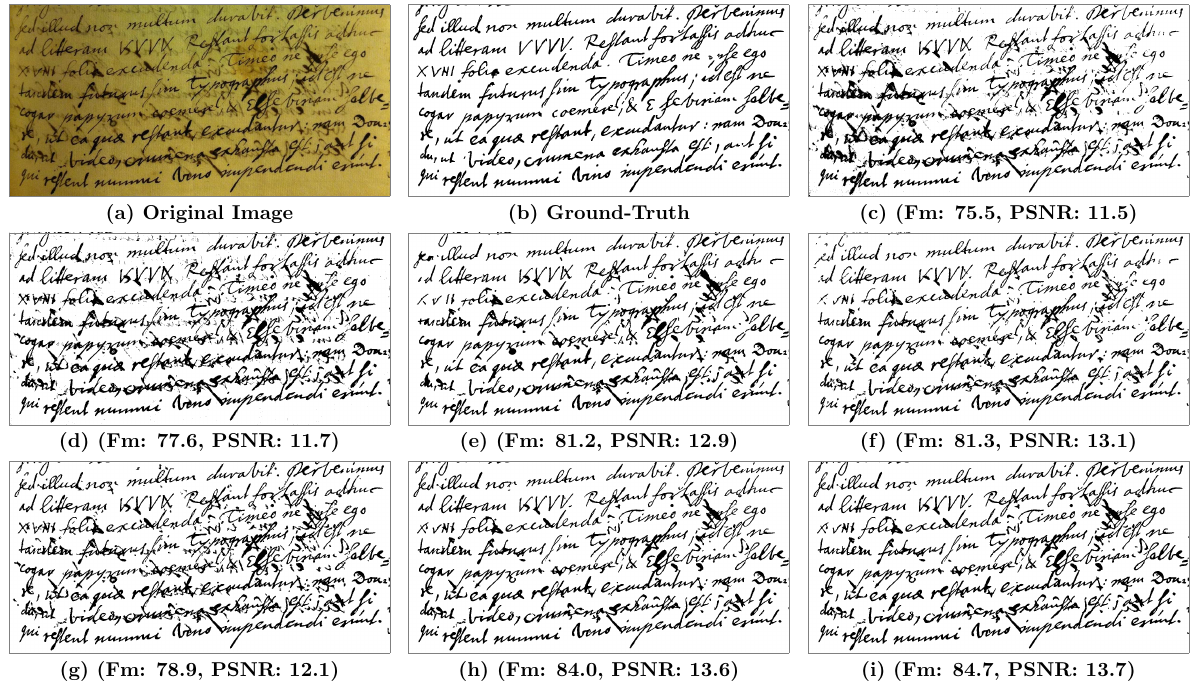}
\caption{Visual example with the F-measure (\textbf{Fm}) and \textbf{PSNR} metric values of the binarization results of one document on the MSC data set produced by different methods: (a) original image, (b) ground truth, (c) Otsu~\cite{otsu1979threshold}, (d) Sauvola~\cite{sauvola2000adaptive}, (e) Howe~\cite{howe2013document}, (f) Su~\cite{su2010binarization}, (g) Jia~\cite{jia2018degraded}, (h) SR-Otsu and (i) SR-Sauvola.}
\label{fig:mcsBinaryExamp}
\end{figure*} 

In order to quantitatively evaluate the enhancement performance, we apply two simple threshold methods to compute the binary maps: global Otsu's threshold~\cite{otsu1979threshold} and local Sauvola's threshold~\cite{sauvola2000adaptive} both on the original and the enhanced images computed by the proposed methods, similar as~\cite{zagoris2017bio}.
Table~\ref{tab:dibco2013enhanced} and~\ref{tab:dibcomscenhanced} show the increased performance of the binarization maps computed on the original images (OI) and the enhanced images by the proposed SR and RR methods on the DIBCO 2013 and MCS datasets, respectively.
From these two tables we can see that the significant improvement is achieved using the proposed deep enhancement methods as a preprocessing step.
Especially, the Sauvola's threshold with the SR methods provides best performance on the two data sets, because the background of the whole image does not uniform since our model works on small patches.

Table~\ref{tab:msccomparsion} shows the performance of different methods on the MCS data set.
The Sauvola's threshold based on the enhanced images produced by the SR method provides the better performance than other traditional methods on this data set.
Fig.~\ref{fig:mcsBinaryExamp} presents the binary results of one document on the MCS data set with different methods.
Our proposed method provides the best visual quality and the values of evaluation metrics (F-measure and PSNR).

\begin{table}[!t]
\centering
\caption{Performance of binarization methods on original and enhanced images on the DIBCO 2013 data set.
Values with red color show the performance improvements of the binarization results on the enhanced images (by SR and RR methods) comparing to the results on the original images (OI).}
\label{tab:dibco2013enhanced}

\begin{tabular}{lcccc}
\hline\hline 
Methods &  F-measure & F$_{ps}$ & PSNR & DRD \\
\hline 
OI-Otsu~\cite{otsu1979threshold} 	& 80.01 & 82.82 & 16.62 & 11.00 \\
SR-Otsu          				& 90.00({\color{red}+9.99}) & 91.68({\color{red}+8.86}) & 20.25({\color{red}+3.63}) & 6.71({\color{red}-4.29}) \\
RR-Otsu          				& 88.90({\color{red}+8.89}) & 91.19({\color{red}+8.37}) & 19.62({\color{red}+3.00}) & 7.07({\color{red}-3.39}) \\
\hline
OI-Sauvola~\cite{sauvola2000adaptive}	& 81.23 & 83.55 & 16.60 & 11.39 \\
SR-Sauvola							& 91.90({\color{red}+10.67}) & 93.79({\color{red}+10.24}) & 20.65({\color{red}+4.05}) & 2.60({\color{red}-8.79})\\
RR-Sauvola							& 90.48({\color{red}+9.35}) & 93.63({\color{red}+10.08}) & 19.97({\color{red}+3.37}) & 2.91({\color{red}-8.48}) \\
\hline\hline  
\end{tabular}
\end{table}

\begin{table}[!t]
\centering
\caption{Performance of other binarization method on original and enhanced images on the MCS data set.
Values with red color show the performance improvements of the binarization results on the enhanced images (by SR and RR methods) comparing to the results on the original images (OI).}
\label{tab:dibcomscenhanced}

\begin{tabular}{lcccc}
\hline\hline 
Methods &  F-measure & F$_{ps}$ & PSNR & DRD \\
\hline 
OI-Otsu~\cite{otsu1979threshold} 	& 69.28 & 70.51 & 11.80 & 33.96\\
SR-Otsu          				& 82.77({\color{red}+13.49}) & 85.80({\color{red}+15.29}) & 15.29({\color{red}+3.49}) & 11.32({\color{red}-22.64})\\
RR-Otsu          				& 79.80({\color{red}+10.52}) & 82.31({\color{red}+11.80}) & 14.54({\color{red}+2.74}) & 15.84({\color{red}-18.12}) \\
\hline
OI-Sauvola~\cite{sauvola2000adaptive}	& 75.84 & 76.85 & 13.08 & 21.54 \\
SR-Sauvola							& 87.01({\color{red}+11.17}) & 89.86({\color{red}+13.01}) & 16.19({\color{red}+3.11}) & 6.07({\color{red}-15.47})\\
RR-Sauvola							& 86.71({\color{red}+10.87})  & 89.68({\color{red}+12.83})  & 16.05({\color{red}+2.97})  & 6.03({\color{red}-15.51})  \\
\hline\hline  
\end{tabular}
\end{table}

\begin{table}
	\centering
	\caption{Comparisons of different algorithms on the MSC data set.}
	\label{tab:msccomparsion}

		\begin{tabular}{lcccc}
			\hline\hline 
			Methods & F-measure & F$_{ps}$ & PSNR & DRD \\
			\hline 
			Otsu~\cite{otsu1979threshold} 			& 69.3 & 70.5 & 11.8 & 34.0 \\
			Sauvola~\cite{sauvola2000adaptive}		& 75.8 & 76.9 & 13.1 & 21.5\\
			Howe~\cite{howe2013document}			& 85.6 & 89.1 & 15.8 & 6.4 \\
			Su~\cite{su2010binarization}			& 82.8 & 87.4 & 15.2 & 16.8 \\
			Jia~\cite{jia2018degraded}				& 85.4 & 88.7 & 15.8 & 7.1\\
			SR-Sauvola								& \textbf{87.0} & \textbf{89.9} & \textbf{16.2} & \textbf{6.1}\\
			\hline\hline
	\end{tabular}
\end{table}

\subsection{Document binarization}
In this section, 
we first describe how to compute the binary maps given the enhanced images.
Then, we present the performance of the document binarization on three benchmark datasets:
DIBCO 2011, H-DIBCO 2014 and H-DIBCO 2016 data sets.

\subsubsection{Binarization}
\begin{figure}
	\centering 
	\includegraphics[width=0.8\textwidth]{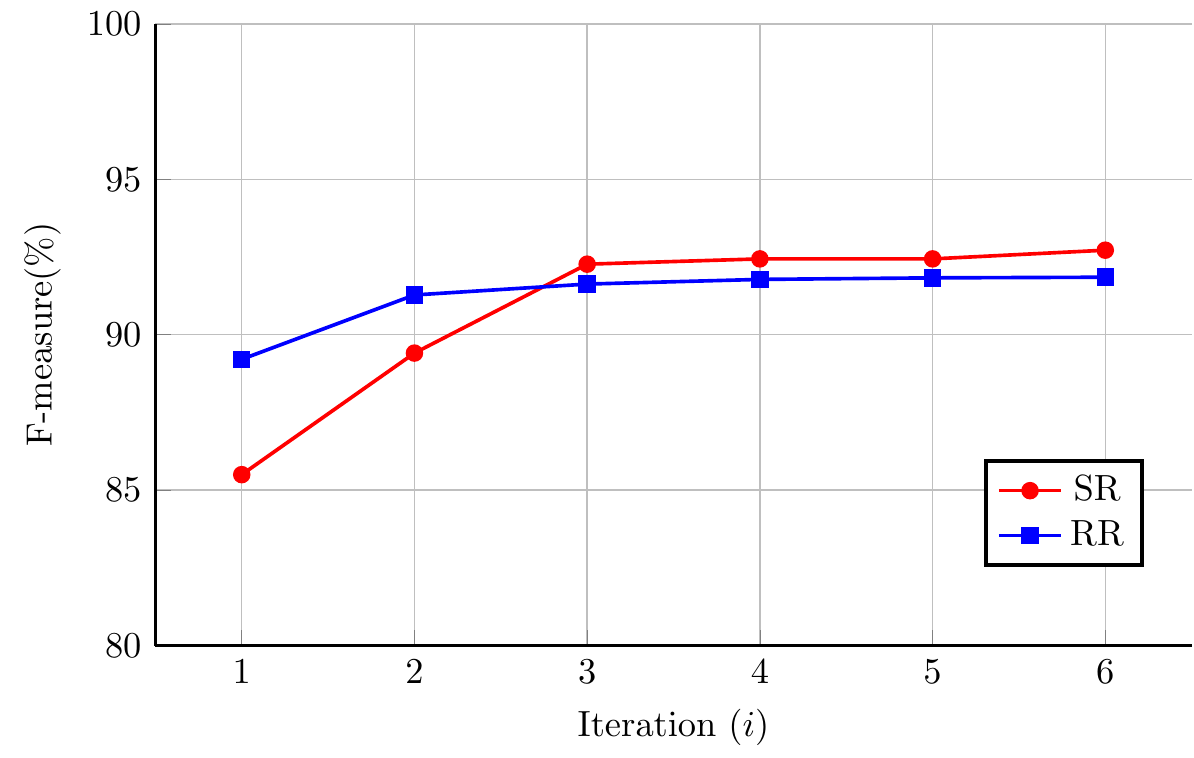}
	\caption{The performance of Otsu's binarization results based on the enhanced images produced by the proposed RR and SR methods on the DIBCO 2011 data set with different iterations $i$ ($i$=1,2,...,6).}
	\label{fig:dibco2011iterations}
\end{figure}

Given the enhanced images produced by the trained neural networks, the existing method can be directly applied to compute the binary maps, such as Otsu~\cite{otsu1979threshold} which is very simple and efficient.
Fig.~\ref{fig:dibco2011iterations} shows the performance of the Otsu's results based on the enhanced images produced by the proposed RR and SR methods with different iterations on the DIBCO 2011 data set.
From the figure we can see that the trained neural network can enhance the document iteratively and the performance is dramatically increasing  at the first three iterations. 
The performance is slightly better with more iterations.

In order to use the global Otsu's threshold, the background of the enhanced images should be uniform.
However, since we use a small patch as input, the enhanced images are not global uniform if the background is nonuniform, which can be found in Fig.~\ref{fig:dibc02013examp}.
To handle this problem, we propose several refinement steps to improve the performance of the Otsu threshold based on the enhancement images by the proposed methods.

\begin{figure*}[!t]
\centering 
\includegraphics[width=\textwidth]{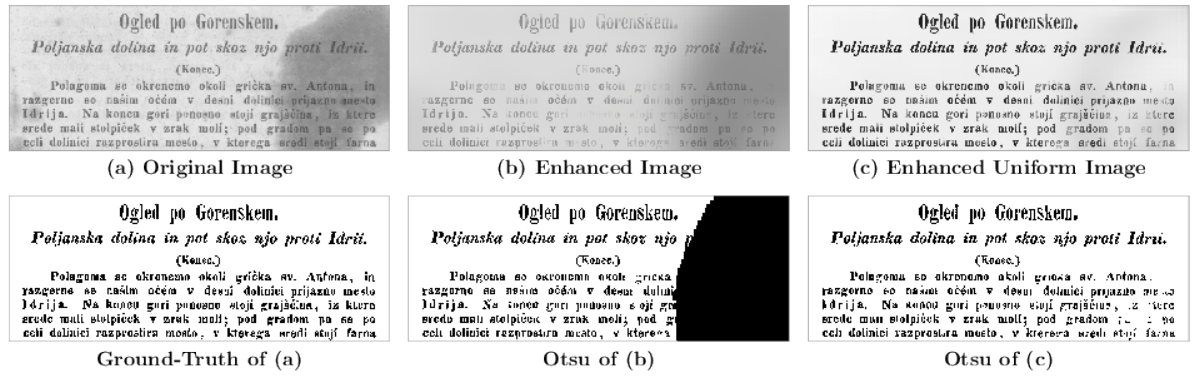}
\caption{Example of locally uniform results of the sample document image (PR04) on the DIBCO 2013 data set.}
\label{fig:uniformExamp}
\end{figure*}

\textbf{Uniform+Otsu}:
We rescale the output of each patch to range of (0,255) if it contains ink trace, which means the background is set to values towards to 255 and the text values towards to 0.
Otherwise we set it to the background.
The Sauvola's threshold is used to determine whether the image patch from the output of the trained neural networks contains the ink trace or not.
Fig.~\ref{fig:uniformExamp} shows an example of the Otsu results with and without using the locally uniform on the document with nonuniform background.
The locally uniform is very helpful to handle the nonuniform background documents when using global Otsu's threshold.

\textbf{MS+Uniform+Otsu}:
We also sample the patches with different scales on the test images (scale factors 0.75, 1.25 and 1.5 based on the size of neural network input) and resize them into 256$\times$256.
The binary map is the average of multiple scale outputs from the neural network.

\textbf{Fusion+MS+Uniform+Otsu}:
Our proposed method can iteratively refine the outputs of the degraded inputs.
However, the weak and thin ink traces might be lost at the end iteration.
The performance can be improved if the outputs of each iteration are integrated.
In this paper, we average the outputs of each iterations (totally six iterations) and then use Otsu to compute the binary maps.

Table~\ref{tab:dibco2011refinement} shows the results on the DIBCO 2011 data set.
From the table we can see that using locally uniform in multiscale can improve the performance of both RR and SR methods.
Combining the outputs from different iterations can provide slightly better performance for the SR method and slightly worse performance in terms of F-measure for the RR method, due to the fact the the SR method contains more convolutional layers than the RR method.
We also compare the results of the Sauvola's threshold~\cite{sauvola2000adaptive} based on the enhanced images computed with different refinements in Table~\ref{tab:dibco2011refinement}.
The local Sauvola's threshold provides a slightly worse performance than the global Otsu's threshold~\cite{otsu1979threshold}.
In the following section, we provide the results of both SR and RR methods using all the refinement steps (Fusion+MS+Uniform+Otsu), named as DeepOtsu.

\begin{table*}[!t]
\centering
\caption{Performance of binarization of the CNN output image with different refinement steps on the DIBCO 2011 data set.}
\label{tab:dibco2011refinement}
\resizebox{\textwidth}{!}{
\begin{tabular}{lcccc|cccc}
\hline\hline 
\multirow{2}{*}{Methods} & \multicolumn{4}{c|}{SR} & \multicolumn{4}{c}{RR}\\
\cline{2-9}
 &  F-measure & F$_{ps}$ & PSNR & DRD &  F-measure & F$_{ps}$ & PSNR & DRD\\
\hline 
Otsu 						& 92.7 & 95.6 & 19.7 & 2.2 & 91.9 & 94.9 & 19.1 & 2.6\\
Uniform+Otsu					& 92.9 & 95.7 & 19.9 & 2.1 & 92.4 & 95.3 & 19.4 & 2.4\\
MS+Uniform+Otsu			& 93.1 & 95.4 & 20.0 & 2.0 & 93.0 & 95.3 & 19.6 & 2.2\\
Fusion+MS+Uniform+Otsu		& 93.4 & 95.8 & 19.9 & 1.9 & 92.8 & 95.6 & 19.5 & 2.2\\
\hline 
Sauvola 					& 90.9 & 93.8 & 19.6 & 2.6 & 90.8 & 94.6 & 18.9 & 2.8 \\
Uniform+Sauvola				& 93.1 & 93.7 & 19.6 & 2.2 & 92.3 & 92.6 & 19.1 & 2.7 \\
MS+Uniform+Sauvola			& 92.2 & 92.2 & 19.1 & 2.5 & 91.9 & 91.7 & 18.7 & 2.7 \\
Fusion+MS+Uniform+Sauvola	& 92.4 & 92.4 & 19.1 & 2.4 & 92.3 & 92.4 & 19.0 & 2.6 \\
\hline\hline
\end{tabular}}
\end{table*}

\subsubsection{Performance on the (H)-DIBCO benchmark data set}
In this section, we provide the comparison performance of the proposed methods with other binarization algorithms on three benchmark data sets.

\begin{figure}[!t]
\centering 
\includegraphics[width=0.8\textwidth]{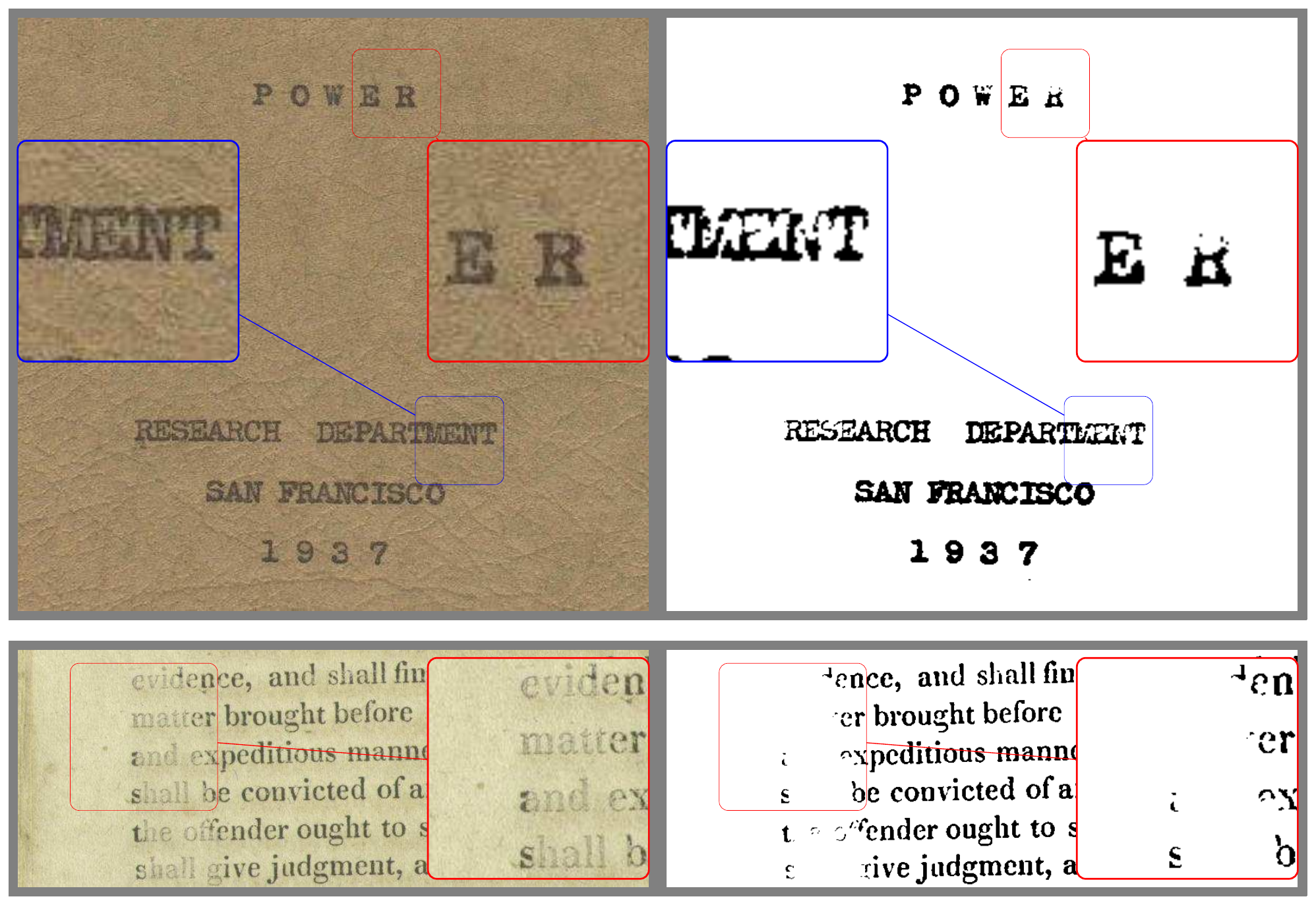}
\caption{Two failure cases on the DIBCO 2011 data set produced by the proposed DeepOtsu (SR) method. The weak strokes are missed in the binary maps.}
\label{fig:dibco2011failures}
\end{figure}

Table~\ref{tab:dibco2012comparsion} shows the performance on the DIBCO 2011 data set.
The Otsu's threshold based on the enhanced images computed by the SR method provides best performance in terms of F-measure and DRD metrics, which shows that our proposed method produces the binarization maps with a less-distorted visual quality.
The method proposed in~\cite{vo2018binarization} also uses a hierarchical neural network to predict the binary maps directly from the degraded images and it provides a slightly better performance in terms of F$_{ps}$ and PSNR.
%The main reason is that the proposed methods consider the thin or weak strokes as the bleed-through ink trace and remove them in the enhanced images.
%This is because the training set contains a large number of bleed-through patches.
Fig.~\ref{fig:dibco2011failures} provides two failure examples on the DIBCO 2011 data set of our proposed method, which misses the weak ink strokes because in the training set, the ground-truth of the weak ink strokes is also very weak which is computed by the average ink pixels in a local patch.
Thus, these text regions are missed in final the binary maps computed by the Otsu's threshold.
This problem could be solved by training the network with more iterations with training samples of weak ink strokes.

\begin{table}[!t]
	\centering
	\caption{Comparisons of different algorithms on the DIBCO 2011 data set.}
	\label{tab:dibco2012comparsion}
%	\resizebox{0.5\textwidth}{!}{
		\begin{tabular}{lcccc}
			\hline\hline 
			Methods & F-measure & F$_{ps}$ & PSNR & DRD \\
			\hline 
			Otsu~\cite{otsu1979threshold} 			& 82.1 & 84.8 & 15.7 & 9.0 \\
			Sauvola~\cite{sauvola2000adaptive}		& 82.1 & 87.7 & 15.6 & 8.5 \\
			Howe~\cite{howe2013document}			& 91.7 & 92.0 & 19.3 & 3.4 \\
			Su~\cite{su2010binarization}			& 87.8 & 90.0 & 17.6 & 4.8 \\
			Jia~\cite{jia2018degraded}				& 91.9 & 95.1 & 19.0 & 2.6 \\
			Vo~\cite{vo2018robust}					& 88.2 & 90.3 & 20.1 & 2.9 \\
			Vo~\cite{vo2018binarization}			& 93.3 & \textbf{96.4} & \textbf{20.1} & 2.0 \\
			DeepOtsu(RR)							& 92.8 & 95.6 & 19.5 & 2.2\\
			DeepOtsu(SR)							& \textbf{93.4} & 95.8 & 19.9 & \textbf{1.9} \\
			\hline\hline
	\end{tabular}
\end{table}

\begin{figure*}[!t]
\centering 
\includegraphics[width=\textwidth]{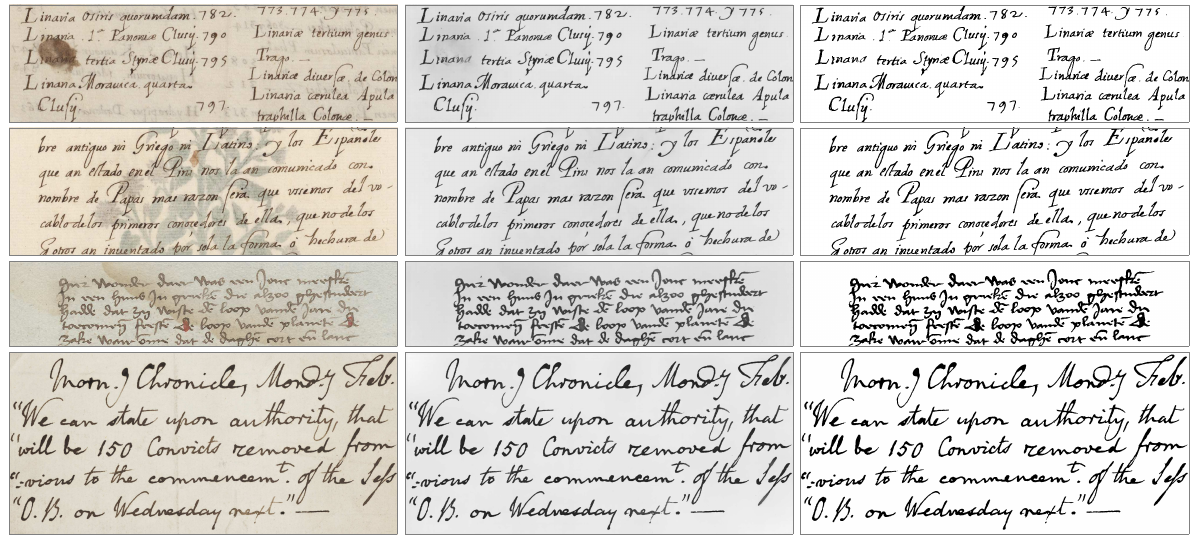}
\caption{Examples of the enhanced and binarization results of sample document images on the H-DIBCO 2014 data set.
The left column shows the original images, the middle column shows the enhanced images by the proposed SR method and the right column shows the binarization maps based on the enhanced images.}
\label{fig:dibco2014exampls}
\end{figure*}

Table~\ref{tab:hdibco2014comparsion} shows the performance of different methods on the historical document competition H-DIBCO 2014 data set.
Fig.~\ref{fig:dibco2014exampls} shows the enhanced and corresponding binarization maps computed by the proposed DeepOtsu (SR) method on this data set.
From Table~\ref{tab:hdibco2014comparsion} we can see that the performance of our proposed DeepOtsu methods is comparable to the Vo's method~\cite{vo2018binarization} which also uses deep learning.
The performance of the traditional methods, such as Howe~\cite{howe2013document}, Su~\cite{su2010binarization} and Jia~\cite{jia2018degraded}, is the comparable to the deep learning methods, such as Vo~\cite{vo2018binarization}, which indicates that the binarization problem on the H-DIBCO 2014 data set is less challenging than other data sets.
The best performance is achieved by the Vo's method~\cite{vo2018binarization}, which integrates outputs of the binary predictions of three different networks and the final binary map is computed by a local and global threshold that is learned on a separate data set.

\begin{table}[!t]
	\centering
	\caption{Comparisons of different algorithms on the H-DIBCO 2014 data set.}
	\label{tab:hdibco2014comparsion}
%	\resizebox{0.5\textwidth}{!}{
		\begin{tabular}{lcccc}
			\hline\hline 
			Methods & F-measure & F$_{ps}$ & PSNR & DRD \\
			\hline 
			Otsu~\cite{otsu1979threshold} 			& 91.7 & 95.7 & 18.7 & 2.7 \\
			Sauvola~\cite{sauvola2000adaptive}		& 84.7 & 87.8 & 17.8 & 2.6 \\
			Howe~\cite{howe2013document}			& 96.5 & 97.4 & 22.2 & 1.1 \\
			Su~\cite{su2010binarization}			& 94.4 & 95.9 & 20.3 & 1.9 \\
			Jia~\cite{jia2018degraded}				& 95.0 & 97.2 & 20.6 & 1.2 \\
			Vo~\cite{vo2018binarization}			& \textbf{96.7} & \textbf{97.6} & \textbf{23.2} & \textbf{0.7} \\
			DeepOtsu(RR)							& 94.3 & 96.3 & 20.9 & 1.9 \\
			DeepOtsu(SR)							& 95.9 & 97.2 & 22.1 & 0.9 \\
			\hline\hline
	\end{tabular}
\end{table}

\begin{figure*}[!t]
\centering 
\includegraphics[width=\textwidth]{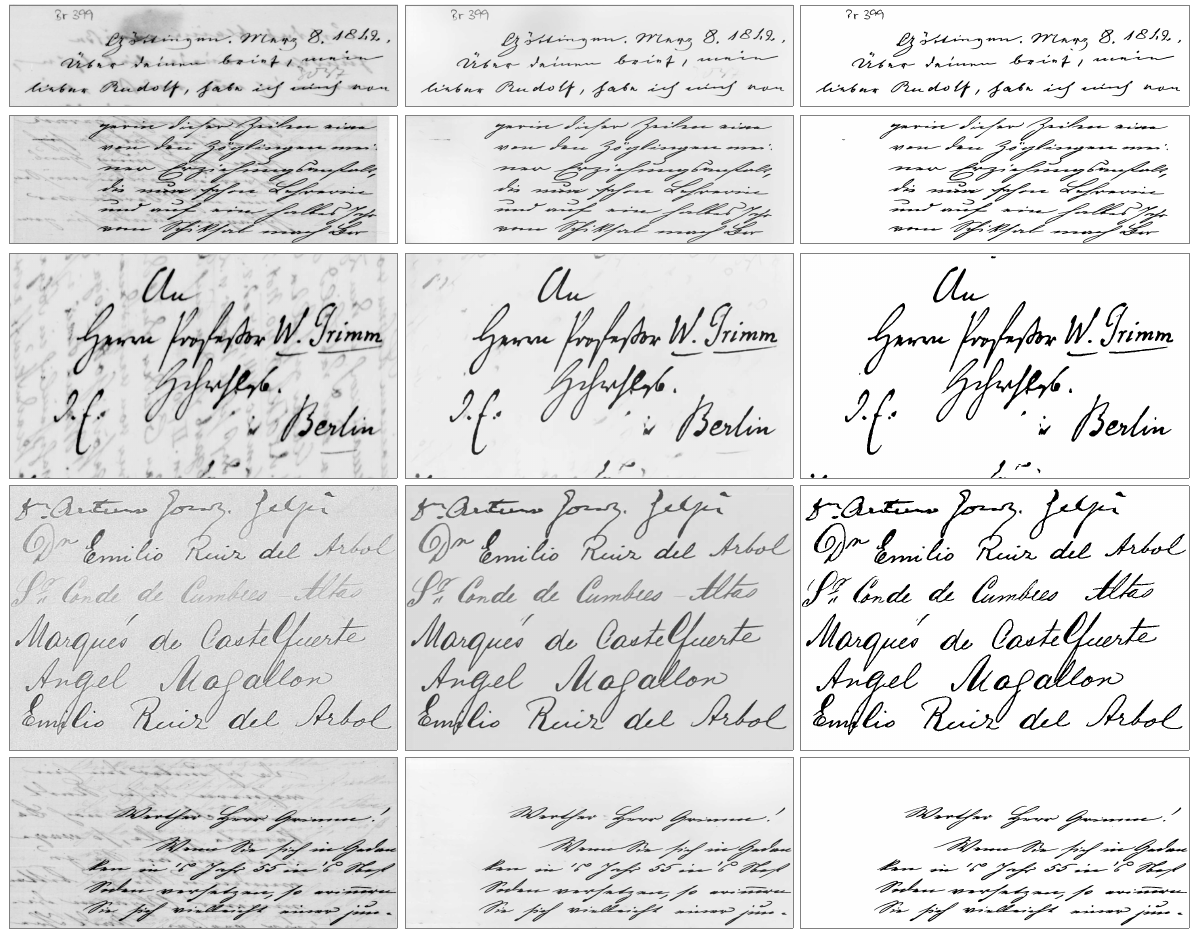}
\caption{Examples of the enhanced and binarization results of sample document images on the H-DIBCO 2016 data set.
The left column shows the original images, the middle column shows the enhanced images by the proposed SR method and the right column shows the binarization maps based on the enhanced images.}
\label{fig:dibco2016exampls}
\end{figure*}

Table~\ref{tab:hdibco2016comparsion} presents the performance of different binarization methods on the H-DIBCO 2016 data set.
Our proposed method provides better results than the traditional threshold methods and also than other deep learning methods, such as the recurrent neural network~\cite{westphal2018document} and the hierarchical deep supervised network~\cite{vo2018binarization}.
Fig.~\ref{fig:dibco2016exampls} shows the enhanced and corresponding binarization maps computed by the proposed DeepOtsu (SR) method on this data set and we can see that the enhanced images are uniform and clean without noise and textures on the background.

\begin{table}[!t]
	\centering
	\caption{Comparisons of different algorithms on the H-DIBCO 2016 data set.}
	\label{tab:hdibco2016comparsion}
%	\resizebox{0.5\textwidth}{!}{
		\begin{tabular}{lcccc}
			\hline\hline 
			Methods & F-measure & F$_{ps}$ & PSNR & DRD \\
			\hline 
			Otsu~\cite{otsu1979threshold} 			& 86.6 & 89.9 & 17.8 & 5.6 \\
			Sauvola~\cite{sauvola2000adaptive}		& 84.6 & 88.4 & 17.1 & 6.3 \\
			Howe~\cite{howe2013document}			& 87.5 & 92.3 & 18.1 & 5.4 \\
			Su~\cite{su2010binarization}			& 84.8 & 88.9 & 17.6 & 5.6 \\
			Jia~\cite{jia2018degraded}				& 90.5 & 93.3 & 19.3 & 3.9 \\
			Vo~\cite{vo2018robust}					& 87.3 & 90.5 & 17.5 & 4.4 \\
			Vo~\cite{vo2018binarization}			& 90.1 & 93.6 & 19.0 & 3.5 \\
			Westphal~\cite{westphal2018document}	& 88.8 & 92.5 & 18.4 & 3.9 \\
			DeepOtsu(RR)							& 90.9 & 93.9 & 19.4 & 3.1 \\
			DeepOtsu(SR)							& \textbf{91.4} & \textbf{94.3} & \textbf{19.6} & \textbf{2.9} \\
			\hline\hline
	\end{tabular}
\end{table}
	
From the above analysis, the proposed method works much better on document images which contains the degradation of bleed-through and noise. 
It provides an improved version of the degraded documents for visualization.
Since the input of our method is a small patch, the large smears can not be removed completely ( Fig.~\ref{fig:dibc02013examp}).
But this problem can be solved by training a neural network with a large input patch.
The trained neural network focuses on removing the degradations, 
the risk is that thin or weak strokes are considered as the degradations by the neural network (Fig.~\ref{fig:dibco2011failures}).
If reconstruction of thin strokes would be needed, this must be reflected in the composition of the training set, containing a sufficient number of such patterns.

	\subsubsection{Computing time analysis}
	The input patch size of the proposed is fixed, which can be computed in a very efficient way by GPU. The training takes about 24 hours for both SR and RR methods on a single GPU (NVIDIA GTX 960 with 4G memory).
	For testing, the computing time of the neural network for each patch is around 0.02 seconds and the processing uniform for binarization takes around 0.0008 seconds.
	The patches on one image can be processed in parallel on different GPUs.
	The training and testing time can be reduced with a stronger computer system with more GPUs and memory.

	\section{Conclusion}
	\label{sec:conclusion}
	
	We have proposed a novel model for document enhancement and binarization based on iterative deep learning. 
	Given a small patch sampled from image, the uniform image is predicted iteratively by the proposed enhancement model in two possible ways: recurrent refinement and stacked refinement.
	The enhanced image produced by the proposed method is a very good view for end users, which is clean, locally uniform and does not contain any undescribable textures in the background.
	We evaluate the proposed method on a real historical collections from the Monk system and several public benchmark data sets.
	The experimental results demonstrate that our method achieves a promising performance.
	
	In this paper, we have used the basic U-Net neural network to learn degradations in document images.
	More complicated neural networks can be adopted in future work, such as the ResNet~\cite{he2016deep} and DenseNet~\cite{huang2017densely} and DSN used in~\cite{vo2018binarization}.
	In addition, the networks in different iterations can also be different.
	The complicated neural network can be applied in the beginning iterations and light neural network can be used in the end iterations for fine-tunning.

	\section*{Acknowledgments}
	
	This work has been supported by the Dutch Organization for Scientific Research NWO Digging into data grant 'Global Currents' (Project no. 640.006.015). The authors would like to thank Zhenwei Shi to label the MCS data set.
	
	\bibliography{BinaryPR}
	
\end{document}